\pdfoutput=1
\documentclass[11pt]{article}
\usepackage{acl} %\usepackage[review]{acl}
\usepackage{times}
\usepackage{latexsym}
\usepackage{tikz}
\usepackage{wrapfig}
\usepackage{float}
\usepackage{amsmath}
\usepackage{listings}
\usepackage{xcolor}
\usepackage{booktabs}
\usepackage[T1]{fontenc}
\usepackage[utf8]{inputenc}
\usepackage{microtype}
\usepackage{graphicx}
\usepackage{placeins}
\lstdefinestyle{schema}{
  basicstyle=\ttfamily\footnotesize,
  breaklines=true,
  columns=fullflexible,
  keepspaces=true,
  showstringspaces=false,
  frame=single
}

\title{ARGUS: Role-Aware Event Knowledge Graphs \\
for U.S. Employment-Discrimination Complaints}

\author{
Sriram Kannan\textsuperscript{1} \qquad
Swetha Saseendran\textsuperscript{1} \qquad
Vishnu Vardhan Reddy Kandi\textsuperscript{1} \\
Leslie Barrett\textsuperscript{2} \qquad
Madhavan Seshadri\textsuperscript{2} \qquad
Enrico Santus\textsuperscript{2} \\
\textsuperscript{1}University of Massachusetts Amherst \\
\textsuperscript{2}Bloomberg \\
\texttt{\{sriramkannan, ssaseendran, vkandi\}@umass.edu} \\
\texttt{\{lbarrett4, mseshadri, esantus\}@bloomberg.net}
}

\begin{document}
\maketitle

\begin{abstract}
U.S.\ employment-discrimination complaints describe complex event sequences that are not explicitly captured by lexical or embedding-based representations alone. We present ARGUS, a source-grounded pipeline that combines a 5W1H-inspired schema, legal-domain models, and LLM-based structured generation to construct document-level Event Knowledge Graphs (EKGs) from CourtListener complaints. ARGUS extracts fact-bearing statements, builds chunk-level event graphs with participant, temporal, and causal structure, and merges them into document-level representations. We evaluate graph quality through human and multi-model assessment and test downstream utility on claim classification and legal QA. The graph-structured classifier outperforms raw and linearized baselines on the held-out set, and EKG-only retrieval improves document-scoped QA, while open-retrieval gains remain limited by low first-stage candidate recall. These results suggest that EKGs are most useful for organizing and reasoning over evidence once relevant material has been retrieved.
\end{abstract}

\section{Introduction}

Legal practitioners often need to interpret and compare cases whose
underlying events are similar even when described differently. Lexical and embedding-based retrieval methods can identify semantically relevant passages, but return largely unstructured text and do not explicitly represent \emph{who did what to whom, when, and why}. This structure is important in legal narratives, where participant roles and temporal or causal relations can determine whether two cases are meaningfully comparable.

We study this problem in U.S.\ employment-discrimination complaints. We present ARGUS, a pipeline that converts complaint narratives into source-grounded Event Knowledge Graphs (EKGs). Rather than representing a case only as text or a set of entities, ARGUS models events together with their participants, legal roles, source evidence, and supported temporal and causal relations.

\begin{figure}[t]
    \centering
    \includegraphics[width=\columnwidth]{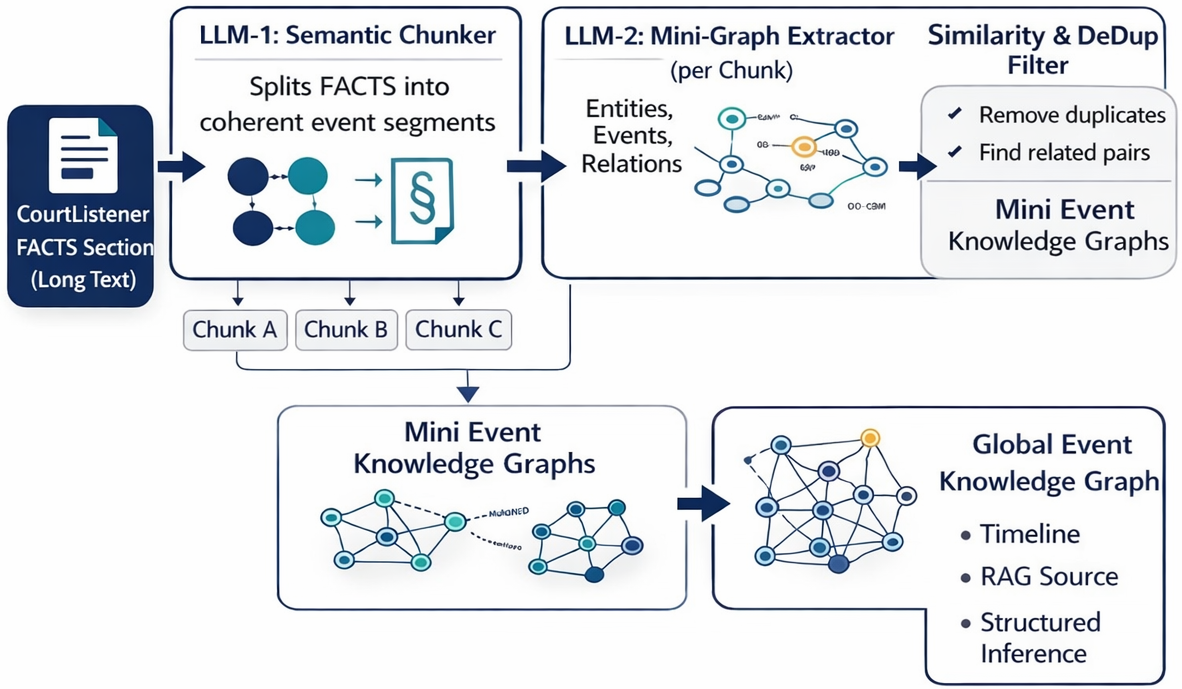}
    \caption{Multi-stage pipeline for EKG construction from federal employment-discrimination complaint narratives.}
    \label{fig:event-pipeline}
\end{figure}

ARGUS follows a staged construction process. Starting from CourtListener complaints,\footnote{\url{https://www.courtlistener.com/}} it identifies fact-bearing statements, groups them into semantic chunks, extracts chunk-level event graphs using a 5W1H-inspired representation \citep{hamborg2019giveme5w1h}, and merges them into a document-level EKG through deterministic and LLM-assisted cross-chunk resolution. Figure~\ref{fig:event-pipeline} provides an overview; implementation and evaluation details are given in Section~\ref{sec:method} and the Appendix.

We investigate three questions: \textbf{(i)} Can the staged pipeline produce reliable, source-grounded legal event graphs? \textbf{(ii)} Do these graphs improve legal classification and document-scoped question answering relative to text-based representations? \textbf{(iii)} Do these benefits extend to external and open-retrieval legal QA settings?

We evaluate graph construction at multiple stages, including fact extraction, chunk-level graph extraction, and document-level merging, using automatic measures, human annotation, and multi-model judging. We then assess downstream utility through claim-type classification and document-scoped FAQ question answering, followed by bar-exam experiments testing generalization to multiple-choice QA and open retrieval. Exploratory case-clustering analyses are reported in Appendix~\ref{app:clustering-details}.

Our contributions are threefold. \textbf{(1)} We introduce a source-grounded, role-aware EKG pipeline for U.S.\ employment-discrimination complaints that represents events with participant, temporal, and causal structure. \textbf{(2)} We introduce a multi-stage evaluation protocol for legal event graphs, combining human annotation with complementary automatic and LLM-based evaluation.\footnote{The human annotations will be released upon publication.} \textbf{(3)} We show that the graph-structured classifier outperforms raw and linearized baselines on held-out claim classification and that EKG-based retrieval improves document-scoped QA; in open retrieval, gains remain limited when relevant evidence is absent from the first-stage candidate set.

%The methodological contribution lies in the staged, source-grounded construction and evaluation of document-level legal event graphs; the downstream experiments test whether these structured representations provide utility beyond text-based representations.

\section{Related Work}

ARGUS intersects three lines of work: legal retrieval and representation, event extraction and temporal reasoning, and graph-augmented retrieval.

\paragraph{Legal retrieval and representation.}
Legal case retrieval has progressed from lexical matching toward neural interaction and representation learning \citep{feng-etal-2024-legal}. BERT-PLI models paragraph-level interactions between cases \citep{shao2020bertpli}, while legal-domain models such as LEGAL-BERT \citep{chalkidis-etal-2020-legal} and models trained on LeXFiles \citep{chalkidis-etal-2023-lexfiles} provide domain-adapted representations for legal NLP. The reasoning-focused legal retrieval benchmark used in our bar-exam experiments separates retrieval quality from downstream answer selection \citep{zheng2025legalretrieval}.

\paragraph{Event extraction and temporal structure.}
General information-extraction systems such as DyGIE++ \citep{wadden-etal-2019-entity}, MetaGraph \citep{pedinotti-etal-2026-metagraph} and OneIE \citep{lin-etal-2020-joint} jointly model entities, relations, and/or events. In the legal domain, \citet{filtz2020events} extract event types, participants, and temporal information from ECHR decisions, while LexTime evaluates temporal ordering of event pairs in U.S.\ federal complaints \citep{barale-etal-2025-lextime}. The 5W1H framework provides a compact representation of core event information \citep{hamborg2019giveme5w1h}. REGen combines exact, relaxed, and LLM-based matching for generative event-argument evaluation and validates these measures against human judgments \citep{sharif-etal-2025-regen}. ARGUS applies a 5W1H-inspired schema to source-grounded mini-graphs and merges them into document-level EKGs with legal participant roles and temporal and causal relations.

\paragraph{Legal knowledge graphs and graph-augmented retrieval.}
\citet{li2024legalkg} construct a Chinese legal knowledge graph from criminal-law materials using a knowledge-enhanced LLM. More broadly, GraphRAG constructs entity-based graph representations for query-focused summarization \citep{edge2024graphrag}, while KG$^2$RAG uses knowledge-graph relations to expand and organize retrieved chunks \citep{zhu-etal-2025-knowledge}. EventRAG builds event knowledge graphs for multi-event reasoning \citep{yang-etal-2025-eventrag}. In the legal domain, \citet{kordjamshidi-etal-2026-reasoners} investigate structured representations and their efficacy in tax-law reasoning for QA, and LegalGraphRAG \citep{chen-etal-2026-legalgraphrag} combines hierarchical legal graphs with multi-agent retrieval and verification. ARGUS instead focuses on source-grounded, document-level EKGs for U.S.\ employment-discrimination complaints, preserving legal participant roles and temporal and causal relations through staged document-level merging.

\section{Method}
\label{sec:method}

ARGUS converts the FACTS narrative of a U.S.\ employment-discrimination complaint into a source-grounded, document-level Event Knowledge Graph (EKG). The pipeline has four stages: \textbf{(1)} sentence-level fact extraction, \textbf{(2)} semantic chunking, \textbf{(3)} two-stage mini-graph extraction, and \textbf{(4)} deterministic and LLM-assisted document-level merging. Figure~\ref{fig:event-pipeline} summarizes the pipeline. We evaluate each component separately to distinguish errors introduced during fact selection, event and relation extraction, and graph merging.

\paragraph{Fact extraction.}
We define facts as event-bearing statements describing concrete actions, communications, decisions, or outcomes, distinguishing them from procedural or rhetorical text. Each sentence $s$ is classified as \emph{Fact} or \emph{Non-Fact} using a fine-tuned LEGAL-BERT or LexLM encoder \citep{chalkidis-etal-2020-legal,chalkidis-etal-2023-lexfiles}. Let $p(s)$ denote the classifier's maximum softmax probability. Predictions above a confidence threshold $\tau$ are retained directly; sentences with $p(s)<\tau$ are passed to an LLM, which returns the identifiers of sentences judged to be facts. Only sentences ultimately labeled as facts are passed to graph construction.

% SUGGETION: Reconcile the confidence threshold before submission. The
% manuscript previously reports \tau=0.70, whereas the uploaded inference
% and evaluation scripts use \tau=0.80. Also identify the encoder and LLM
% refiner used to generate the fact labels underlying the downstream EKGs.

\paragraph{Semantic chunking.}
Retained fact sentences preserve their original identifiers and are formatted as sentence-labeled input (\texttt{[S1]}, \texttt{[S2]}, \ldots). An LLM groups them into coherent event sequences, topics, or narrative phases. The prompt favors boundaries associated with changes in theme, time, or actor, avoids splitting an event across chunks, requires coverage of all retained sentences, and introduces approximately two to three sentences of overlap.

% SUGGETION: State the exact chunking model and fallback configuration used
% in the reported run. The implementation applies LLM chunking only below
% a configured sentence limit and otherwise uses a sentence-based fallback.

\paragraph{Event representation.}
ARGUS uses a 5W1H-inspired representation \citep{hamborg2019giveme5w1h} within a structured event-extraction formulation \citep{srivastava-etal-2025-instruction}. Entity records contain an identifier, name, entity kind, canonical legal role, and optional aliases. Event records contain an identifier, event type, main verb, trigger, participants, temporal information, source evidence, and confidence. Participants reference entity identifiers and carry event-specific roles, while evidence retains supporting sentence identifiers.

Temporal information may be explicit, relative, or unknown, following the broader problem of event-pair temporal relation modeling \citep{ning-etal-2018-multi}. The extraction schema supports \textsc{before}, \textsc{after}, \textsc{overlap}, and \textsc{same-time} temporal relations and \textsc{causes}, \textsc{enables}, and \textsc{prevents} causal relations. ARGUS retains relations only when supported by the source text. For graph-based classification, events and participant entities become nodes; event--entity edges represent participation, and temporal and causal relations remain event--event edges. An example extracted mini-graph is shown in Appendix~\ref{app:pipeline-figures}, Figure~\ref{fig:minigraph-example}; the full downstream GNN architecture is provided in Appendix~\ref{app:gnn-details}.

% SUGGETION: Harmonize the relation inventory across the Introduction,
% Method, baseline prompt, and extraction code. Also remove polarity and
% location from the paper unless they occur in the final extraction schema.

\paragraph{Two-stage mini-graph extraction.}
Each chunk is processed through two structured-generation stages. Stage~1 extracts grounded entities and events and is explicitly instructed not to produce temporal or causal edges. Stage~2 receives the original chunk and the complete Stage~1 JSON output and adds only source-supported temporal and causal relations. Both stages require strict JSON, preservation of sentence identifiers, and no unsupported entities, events, dates, or relations.

The extraction configuration is optimized with GEPA \citep{agrawal2025gepa}. A candidate specifies the Stage~1 and Stage~2 models, prompts, retry instructions, and extraction settings. Candidates are evaluated using structural checks---including schema validity, reference integrity, temporal and causal consistency, and coverage---together with an LLM judge that scores factual accuracy, completeness, relevance, faithfulness, coherence, and overall quality. Cost is measured using token counts when available and API-call count otherwise. GEPA uses the metrics and observed failures to propose mutations and retains non-dominated configurations on a quality--cost Pareto frontier. The two-stage extractor and GEPA optimization loop are shown in Appendix Figures~\ref{fig:minigraph-extractor} and~\ref{fig:gepa-loop}.

We also evaluate a Generative FrameNet-guided configuration \citep{tayyar-madabushi-etal-2025-generative}, using frame hints such as hiring, retaliation, and termination to guide event typing.

% SUGGETION: Identify the final optimized candidate used to generate the
% downstream EKGs. The final candidate and FrameNet-guided prompt should
% also be included in the released code or Appendix.

\paragraph{Document-level graph merging.}
Mini-graphs are first combined through deterministic merging. Entity identifiers are remapped into a document-level namespace, and mentions with the same lowercased, whitespace-normalized name are assigned to the same entity. Events are deduplicated using a structured key containing event type, normalized main verb, temporal value, participant entity--role pairs, and supporting sentence identifiers. Duplicate records accumulate their source references, and existing intra-chunk edges are remapped to the resulting document-level event identifiers.

% SUGGETION: The uploaded implementation deduplicates entities by normalized
% exact name. It does not implement the TF-IDF and type/role compatibility
% procedure described in the previous paper draft. Align the implementation
% and manuscript before submission.

ARGUS then performs LLM-assisted cross-chunk resolution. Candidate pairs are restricted to events originating from different chunks and ranked by similarity over their event descriptions and contextual evidence. For each candidate, the LLM determines whether the records describe the same occurrence and whether a supported temporal or causal relation should be added. The prompt favors precision and returns no relation when evidence is insufficient. Duplicate events may be collapsed, after which existing edges are remapped and accepted cross-chunk relations are added.

% SUGGETION: Identify the exact retrieval implementation used for the
% reported merger experiment. The repository contains both direct embedding
% retrieval and a newer lexical/structural procedure with semantic rescue.
% Report only the corresponding model, threshold, top-k, and candidate limits.
%
% SUGGETION: The code does not expose a separate knowledge-expansion stage;
% cross-chunk relation addition occurs within the hybrid merger. Remove
% knowledge expansion as a separate stage unless additional code exists.

\paragraph{Component-level evaluation.}
Fact extraction is evaluated by comparing LEGAL-BERT and LexLM with a prompted GPT-5 baseline, using Fact-class F1 as the primary metric and accuracy, precision, and recall as supporting measures. Mini-graph extraction is evaluated on a fixed 24-chunk set using automatic structural and semantic measures and human annotation. Four annotators inspect each graph with its source text and assign \emph{Aligned}, \emph{Partially Aligned}, or \emph{Misaligned}, together with an error tag and explanation when applicable. We report percent agreement, Fleiss' $\kappa$, and Krippendorff's $\alpha$.

Document-level construction is compared with a one-shot Claude Sonnet~4.5 baseline that generates a complete EKG directly from the full FACTS narrative; the exact prompt is provided in Appendix~\ref{app:oneshot-prompt}. Three LLM judges---DeepSeek-R1, Qwen3-Coder, and Mistral-Large---compare the staged and one-shot graphs on event grounding, evidence grounding, edge correctness, and overall quality. We additionally report inter-judge agreement, Gwet's AC1 \citep{gwet2008computing}, and stability under temperature variation (Appendix~\ref{app:intra-stability}, Table~\ref{tab:intra_model}). Multiple judges complement human annotation and reduce reliance on a single evaluator \citep{zheng2023judging}. U.S.\ holdout results are reported in Appendix~\ref{app:fact-extraction}, Table~\ref{tab:us_holdout}.

% SUGGETION: Reconcile the U.S. fact-extraction split before submission.
% Once confirmed, report the split consistently here and in
% Appendix~\ref{app:fact-extraction}.

\section{Experimental Setup}

\subsection{Datasets}

We collect 153 employment-discrimination complaints (Nature of Suit 442, Civil Employment Cases) from CourtListener,\footnote{\url{https://www.courtlistener.com/}} an open legal-search repository maintained by the Free Law Project.

A complaint is the initiating pleading filed by a plaintiff and describes allegations rather than facts established by a court. Accordingly, the events represented by ARGUS reflect the allegations contained in the complaint and should not be interpreted as adjudicated findings.

The raw filings are OCR-derived scanned court documents with no gold labels for facts, events, or case similarity. We retrieve them through a Selenium-based pipeline and preprocess them using OCR-to-text conversion, header/footer and boilerplate removal, section detection, sentence segmentation, and normalization. Section detection isolates the FACTS narrative from jurisdictional and prayer-for-relief text.

The corpus covers five U.S.\ district courts to include jurisdiction-specific writing variation (Table~\ref{tab:dataset-stats}): Eastern District of Pennsylvania (EAPD), Eastern District of Kentucky (KYED), Southern District of New York (NYSD), Western District of Pennsylvania (PAWD), and District of Rhode Island (RID). Documents vary substantially in length and structure. Where sentence-level annotation is complete, the fact rate ranges from 81.7\% (KYED) to 93.0\% (PAWD). All splits are defined at the document level to prevent boilerplate and near-duplicate leakage; sentence-level fact extraction inherits its parent document split, while mini-graph extraction and merging are inference-only.

\begin{table}[t]
\centering
%\small
\resizebox{\columnwidth}{!}{%
\begin{tabular}{lccccc}
\toprule
\textbf{District} & \textbf{Docs} & \textbf{Chunks} & \textbf{Avg/Doc} & \textbf{Sentences} & \textbf{Fact Rate} \\
\midrule
EAPD & 20 & 87  & 4.35 & --      & --     \\
KYED & 31 & 138 & 4.45 & 1{,}155 & 81.7\% \\
NYSD & 46 & 318 & 6.91 & 4{,}808 & 92.7\% \\
PAWD & 28 & 165 & 5.89 & 1{,}607 & 93.0\% \\
RID  & 28 & 118 & 4.21 & 1{,}631 & 90.9\% \\
\bottomrule
\end{tabular}}
\caption{Dataset statistics across districts. Fact Rate is the fraction of sentences labeled Fact; sentence-level annotation is not yet complete for EAPD.}
\label{tab:dataset-stats}
\end{table}

\subsection{Evaluation of Graph Construction}

\paragraph{Fact extraction.}
We fine-tune LEGAL-BERT and LexLM for sentence-level Fact/Non-Fact classification and compare them with a prompted GPT-5 baseline on a fixed 100-sentence U.S.\ gold sample (Table~\ref{tab:fact-us}). Fact-class F1 is the primary metric, with accuracy, precision, and recall also reported. LEGAL-BERT achieves the highest F1 (0.7273) and precision (0.8889), while GPT-5 achieves substantially higher recall (0.9231). This precision--recall complementarity motivates the hybrid fallback described in Section~\ref{sec:method}. On the U.S.\ holdout, classifier-only and hybrid modes perform similarly (F1 $\approx0.87$ for LEGAL-BERT and $\approx0.86$ for LexLM), whereas LLM-only extraction is weaker (F1 $\approx0.71$--$0.72$; see Appendix~\ref{app:fact-extraction}, Table~\ref{tab:us_holdout}).

% SUGGETION: Confirm the U.S. holdout split corresponding to these results
% and report it consistently here and in Appendix~\ref{app:fact-extraction}.

\begin{table}[t]
\centering
\setlength{\tabcolsep}{4pt}
\begin{tabular}{lcccc}
\toprule
\textbf{Model} & \textbf{Acc.} & \textbf{Prec.} & \textbf{Rec.} & \textbf{F1} \\
\midrule
LEGAL-BERT & 0.9400 & 0.8889 & 0.6154 & 0.7273 \\
LexLM     & 0.9300 & 0.8000 & 0.6154 & 0.6957 \\
GPT-5     & 0.8500 & 0.4615 & 0.9231 & 0.6154 \\
\bottomrule
\end{tabular}
\caption{Fact-extraction results on a fixed 100-sentence U.S.\ gold sample. LEGAL-BERT and LexLM are fine-tuned legal-domain encoders; GPT-5 is a prompted baseline. Fact is the positive class.}
\label{tab:fact-us}
\end{table}

\paragraph{Mini-graph extraction.}
We evaluate mini-graph extraction on a fixed 24-chunk set. Claude Sonnet~4.5 + GEPA achieves a quality score of 0.896, with schema validity and coverage of 1.0; causal self-loops are the main residual structural error. Among open-source alternatives (Table~\ref{tab:open_source_minigraph}), Qwen achieves higher semantic quality and produces more events and relations per chunk, while GPT-OSS-120B scores slightly higher on structural and temporal-consistency measures.

For human evaluation, four team members independently review each mini-graph with its source text and assign \emph{Aligned}, \emph{Partially Aligned}, or \emph{Misaligned}, plus a free-text explanation and, when applicable, an error tag such as missing event, wrong role, or hallucinated detail. We report percent agreement, Fleiss' $\kappa$, and Krippendorff's $\alpha$ (Table~\ref{tab:human_minigraph_alignment}). The FrameNet-guided Qwen configuration reaches 85.2\% Aligned and Fleiss' $\kappa=0.669$, compared with 78.3\% and $\kappa=0.143$ for the Claude configuration. Because model choice and FrameNet guidance vary together, this is a configuration-level comparison rather than an isolated FrameNet ablation.

\begin{table}[t]
\centering
%\small
\begin{tabular}{@{}lcc@{}}
\toprule
\textbf{Metric} & \textbf{GPT-OSS} & \textbf{Qwen} \\
\midrule
Quality Score          & 0.8272 & 0.8852 \\
Judge Score            & 0.6642 & 0.7966 \\
Structural Score       & 0.9902 & 0.9739 \\
Temporal Consistency   & 0.9118 & 0.7647 \\
\midrule
Events/Chunk            & 7.09 & 7.50 \\
Temporal Edges/Chunk    & 2.15 & 3.12 \\
Causal Edges/Chunk      & 1.00 & 2.29 \\
Schema Validity         & 1.00 & 1.00 \\
\bottomrule
\end{tabular}
\caption{Open-source model comparison for mini-graph extraction.}
\label{tab:open_source_minigraph}
\end{table}

\begin{table}[t]
%%\small
\centering
\begin{tabular}{@{}lcc@{}}
\toprule
\textbf{Metric} & \textbf{Claude} & \textbf{Qwen} \\
\midrule
Aligned             & 78.30\% & 85.20\% \\
Partially Aligned   & 17.40\% & 14.80\% \\
Misaligned          & 4.30\%  & 0.00\% \\
\midrule
Percent Agreement   & 69.60\% & 91.70\% \\
Fleiss' $\kappa$    & 0.143   & 0.669 \\
Krippendorff's $\alpha$ & 0.153 & 0.673 \\
\bottomrule
\end{tabular}
\caption{Human evaluation of mini-graph alignment with the source text.}
\label{tab:human_minigraph_alignment}
\end{table}

\paragraph{Document-level graph merging.}
We first measure how document-level merging changes graph size on five validation documents. Table~\ref{tab:merger-stats} shows that event counts are almost entirely preserved: only one event is removed across the five cases. Entity counts decrease substantially in four cases as repeated mentions are consolidated into document-level entities. This provides a structural check before comparison with a one-shot baseline.

\begin{table}[t]
\centering
%%\small
\setlength{\tabcolsep}{5pt}
\resizebox{\columnwidth}{!}{%
\begin{tabular}{lccccc}
\toprule
\textbf{Case} & \textbf{Pre-Ev} & \textbf{Post-Ev} & \textbf{Loss} & \textbf{Pre-En} & \textbf{Post-En} \\
\midrule
Miczulski v.\ Alix & 27 & 26 & 1 & 31 & 13 \\
Poole v.\ Ampler & 40 & 40 & 0 & 34 & 17 \\
Sermarini v.\ Step Up & 35 & 35 & 0 & 26 & 10 \\
Bleiler v.\ Chester & 27 & 27 & 0 & 24 & 13 \\
Bacon v.\ M.A.G. & 5 & 5 & 0 & 6 & 6 \\
\bottomrule
\end{tabular}}
\caption{Event and entity counts before and after document-level merging on five validation cases.}
\label{tab:merger-stats}
\end{table}

We then compare ARGUS with a one-shot baseline in which Claude Sonnet~4.5, accessed through an OpenAI-compatible hosted API, generates a complete document-level EKG directly from the full FACTS section, without semantic chunking, intermediate mini-graphs, or staged merging. The exact prompt is provided in Appendix~\ref{app:oneshot-prompt}. The baseline extracts no events for two of five validation documents and varies substantially in coverage on the others (Table~\ref{tab:baseline_onshot}), demonstrating the difficulty of performing event extraction and relational reasoning over the complete narrative in one generation.

\begin{table}[t]
\centering
\small
\begin{tabular}{lccc}
\toprule
\textbf{Case} & \textbf{Events} & \textbf{Temporal} & \textbf{Causal} \\
\midrule
Miczulski v.\ Alix Inc & 21 & 19 & 6 \\
Poole v.\ Ampler Pizza & 0 & 0 & 0 \\
Sermarini v.\ A Step Up & 16 & 14 & 20 \\
Bleiler v.\ Chester Co. & 0 & 0 & 0 \\
Bacon v.\ M.A.G. Ent. & 17 & 16 & 13 \\
\bottomrule
\end{tabular}
\caption{One-shot LLM baseline event-graph extraction on the five merger-validation documents.}
\label{tab:baseline_onshot}
\end{table}

Three LLM judges---DeepSeek-R1, Qwen3-Coder, and Mistral-Large---compare the one-shot and staged graphs on event grounding, evidence grounding, edge correctness, and overall quality. ARGUS is preferred on evidence grounding, edge correctness, and overall quality in all comparisons and on event grounding in 93.3\% of comparisons (all $p=0.031$; Table~\ref{tab:inter_model}). Intra-model stability under temperature variation is 100\% for Qwen and Mistral and 80\% for DeepSeek-R1 (Appendix~\ref{app:intra-stability}, Table~\ref{tab:intra_model}).

\begin{table}[t]
\centering
\small
\begin{tabular}{lcccc}
\toprule
\textbf{Dimension} & \textbf{Win} & \textbf{Agree.} & \textbf{AC1} & \textbf{$p$} \\
\midrule
Event Grounding     & 93.3\% & 86.7\% & 0.848 & 0.031 \\
Evidence Grounding  & 100\%  & 100\%  & 1.0   & 0.031 \\
Edge Correctness    & 100\%  & 100\%  & 1.0   & 0.031 \\
Overall             & 100\%  & 100\%  & 1.0   & 0.031 \\
\bottomrule
\end{tabular}
\caption{Multi-model evaluation of ARGUS relative to the one-shot baseline.}
\label{tab:inter_model}
\end{table}

\section{Experiments on Downstream Tasks}

\subsection{Classification}

\paragraph{Setup.}
We formulate claim-type prediction as multi-label classification over 23 concepts from the SALI Alliance's Legal Matter Specification Standard (LMSS)\footnote{The LMSS is available at \url{https://sali.org/explore-the-standard/}.} across 130 complaints (176 label assignments), using LLM-generated silver labels reviewed by humans. We compare three representations over the same label space: the preprocessed complaint text encoded from a 512-token input, a linearized entity/event EKG, and a graph-structured EKG using participation, temporal, and causal edges. Raw text provides a text baseline; the linearized EKG tests whether graph topology adds value beyond serialization of the extracted graph.

\paragraph{Results.}
On the 22-document held-out test set (Micro-F1@0.5; Table~\ref{tab:sali_clf}), the raw-text and linearized-EKG LEGAL-BERT models obtain $0.590$ and $0.583$, respectively, while the graph-structured EKG-GNN reaches 0.642. This is an absolute gain of 0.052 Micro-F1, or 8.8\% relative to the raw-text baseline. The graph-structured representation therefore performs best on this held-out set. %Label frequencies are highly imbalanced; the dominant class, \emph{Systemic Discrimination}, appears in 80.77\% of documents.

\begin{table}[t]
\small
\centering
\begin{tabular}{lc}
\toprule
\textbf{Representation} & \textbf{Micro-F1@0.5} \\
\midrule
Raw-text LEGAL-BERT       & 0.5897 \\
Linearized-EKG LEGAL-BERT & 0.5833 \\
\textbf{EKG-GNN}          & \textbf{0.6415} \\
\bottomrule
\end{tabular}
\caption{SALI claim-type classification on the 22-document held-out test set.}
\label{tab:sali_clf}
\end{table}

\paragraph{GNN architecture.}
The EKG-GNN represents participant entities and events as nodes, with event--entity participation edges and event--event temporal and causal edges. Each node's natural-language description is encoded with LEGAL-BERT \citep{chalkidis-etal-2020-legal} and projected to 256 dimensions. Two mean-neighbor graph-convolution layers propagate information over the document graph:
\[
h_i^{(l+1)} =
\mathrm{ReLU}\!\left(
W_s h_i^{(l)}
+
W_n \frac{1}{|\mathcal{N}(i)|}
\sum_{j\in\mathcal{N}(i)} h_j^{(l)}
\right).
\]
Each layer is followed by dropout ($p=0.1$). Self-loops are excluded from the neighbor average because the node's own representation is modeled separately through $W_s$. After two layers, node representations are mean-pooled and passed to a 23-label classification head. The reported \texttt{ft-BERT} configuration jointly fine-tunes LEGAL-BERT and the graph layers using weighted binary cross-entropy and AdamW. Full architecture and training settings are provided in Appendix~\ref{app:gnn-details}.

\subsection{Clustering vs.\ Distance}

\paragraph{Setup.}
Each merged EKG is reduced to Core6D: log-counts of entities, events, temporal edges, and causal edges, plus graph density and temporal-edge ratio. The vectors are L2-normalized and clustered with $K$-means. We use the silhouette score to compare graph-derived representations with a TF-IDF bag-of-words baseline and to evaluate alternative values of $K$.

Core6D reaches a silhouette score of 0.649 at $K=4$, compared with 0.091 for TF-IDF, 0.363 for the full high-dimensional EKG representation, and 0.465 for Legacy7D. Although the $K=2$--$40$ sweep favors very small $K$, we use $K=4$ to retain stronger granularity with high cohesion. Compact structural summaries therefore produce substantially cleaner cluster geometry than TF-IDF or the unreduced EKG representation at this corpus size. Full results are reported in Appendix~\ref{app:clustering-details}, Table~\ref{tab:clustering}.

\paragraph{Feature ablations.}
We compare three Core6D-based configurations using distance to the assigned centroid and an independent LLM-judged cluster-membership score. Run A combines SALI claim tags with Core6D and produces one dominant cluster (111/14/6/2), with Pearson $r \approx-0.06$ between centroid distance and judged membership. Run B augments Core6D with a 16-dimensional PCA-compressed EKG-context embedding, producing more balanced clusters, with Pearson $r \approx-0.28$ and Spearman $\rho \approx-0.31$. Run C adds SALI tags and a $k$NN graph-neighbor block to Run B, reaching Pearson $r \approx-0.29$ and Spearman $\rho \approx-0.37$, without clearly outperforming Run B.

The pattern indicates that continuous EKG-derived context contributes more clustering signal than SALI claim tags alone, while the additional Run C components provide limited further benefit. Validation uses geometric and LLM-judged membership diagnostics rather than attorney relevance judgments. Detailed feature definitions, ablations, and plots are provided in Appendix~\ref{app:clustering-details} (Tables~\ref{tab:cluster-hybrid}--\ref{tab:cluster-features} and Figures~\ref{fig:cluster-fusion}--\ref{fig:cluster-scatter}).

\subsection{Document-Scoped Question Answering}
\label{sec:rag-courtlistener}

\paragraph{Setup.}
We build a 300-question FAQ benchmark over 10 CourtListener employment-discrimination complaints (30 questions per document), spanning four graph-relevant question types: \emph{temporal order} (what occurred before or after an event), \emph{causal/enablement} (what allegedly caused or enabled an event), \emph{temporal overlap} (whether two events were concurrent), and \emph{party resolution} (who did what to whom), plus a small residual temporal-relation category. Each question has a gold answer grounded in its source complaint.

Holding GPT-4o and the answering prompt fixed, we compare three retrieval conditions:

\begin{enumerate}
    \item \textbf{Raw document} --- the FACTS section is chunked and retrieved directly, without graph structure.
    \item \textbf{Hybrid text + EKG} --- linearized document-level EKG context is combined with retrieved raw-text context.
    \item \textbf{EKG only} --- the linearized EKG is used without raw-text context.
\end{enumerate}

Answer quality is measured with Token-F1 against the gold answer. Because each question is paired with its source complaint, this is a within-document retrieval experiment: it evaluates representation and reasoning after the relevant document is known, not first-stage retrieval across CourtListener.

\paragraph{Results.}
EKG-only retrieval achieves 0.446 mean Token-F1, compared with 0.323 for raw text and 0.363 for hybrid text + EKG (Table~\ref{tab:rag-results}), absolute gains of 0.123 and 0.083, respectively. Hybrid improves over raw text by 0.040, but adding raw context reduces performance relative to EKG alone. This pattern is consistent with additional unstructured context introducing noise, although the experiment does not isolate that mechanism.

\begin{table}[t]
\centering
\begin{tabular}{lc}
\toprule
\textbf{Context condition} & \textbf{Mean Token-F1} \\
\midrule
Raw document            & 0.323 \\
Hybrid text + EKG       & 0.363 \\
\textbf{EKG only}       & \textbf{0.446} \\
\bottomrule
\end{tabular}
\caption{Document-scoped FAQ answer quality over 300 questions from 10 CourtListener complaints (GPT-4o).}
\label{tab:rag-results}
\end{table}

\paragraph{Results by question type.}
The EKG gain over raw text is largest for \emph{temporal order} (0.381~$\to$~0.538, $+0.157$) and \emph{causal/enablement} (0.387~$\to$~0.522, $+0.134$), categories directly aligned with graph relations. Temporal-overlap questions also improve substantially (0.241~$\to$~0.384, $+0.143$). The smallest improvement is for party resolution (0.152~$\to$~0.197, $+0.045$), where participant names and mentions are already accessible in source text. Full question-type results are reported in Appendix~\ref{app:faq-significance}, Table~\ref{tab:rag-by-type}.

\paragraph{Significance of RAG gains.}
We test the $n=299$ questions scored under all three conditions; one failed hybrid LLM call is excluded from all three arms to preserve pairing. We use a paired bootstrap 95\% confidence interval over the mean F1 difference (10,000 resamples), Wilcoxon signed-rank test, and sign-flip paired permutation test (10,000 permutations).

All three tests agree in direction and significance. EKG-only exceeds raw text by $+0.123$ F1 ($p<10^{-30}$) and hybrid by $+0.083$ ($p<10^{-20}$), while hybrid exceeds raw text by $+0.040$ ($p<10^{-10}$). Bootstrap confidence intervals and exact Wilcoxon results are reported in Appendix~\ref{app:faq-significance}, Table~\ref{tab:faq-sig}.

Per-type tests show significant EKG-versus-raw improvements for every category with $n\geq12$: temporal order ($p<10^{-17}$), causal/enablement ($p<10^{-15}$), temporal overlap ($p=2.4\times10^{-3}$), and party resolution ($p=8.0\times10^{-5}$). The residual temporal-relation category contains three questions and is not significant ($p=0.25$); full results are reported in Appendix~\ref{app:faq-significance}, Table~\ref{tab:faq-sig-bytype}.

\section{Generalization to External Datasets}
\paragraph{Multiple-Choice Legal QA (Bar Exam).}
\label{sec:mbe-generalization}

To test generalization beyond the employment-complaint FAQ benchmark, we evaluate on Multistate Bar Examination (MBE) multiple-choice questions \citep{reglab2024barexam}. We report two experiments with different retrieval setups and corpus scales; they are evaluated separately.

\subsection{Experiment A: Retrieval-Condition Comparison}
\label{sec:mbe-expA}

\paragraph{Setup.}
We evaluate 117 MBE questions, each containing a fact pattern, four answer choices, and a gold supporting passage drawn from 100 short appellate opinion excerpts. Unlike the FAQ benchmark, the task selects an answer choice (A--D) rather than generating free text, and its passages contain legal holdings and reasoning rather than narrative complaints. We compare raw document, chunked text, and EKG graph retrieval (top 3 passages each), holding Claude Sonnet~5 fixed.

\paragraph{Results.}
Raw document retrieval reaches 85.47\% accuracy, while chunked text and EKG retrieval both reach 86.32\%, a 0.85 percentage-point gain (Table~\ref{tab:mbe-retrieval-accuracy}). Among the 100 questions with an available gold-source EKG, EKG accuracy is 86.00\%.

\paragraph{Significance.}
We use exact McNemar tests, paired bootstrap 95\% confidence intervals (10,000 resamples), and Wilcoxon signed-rank tests. No pairwise comparison is significant: all McNemar $p$-values exceed 0.68 and all bootstrap intervals include zero. Only 6--9 outcomes differ between any pair of conditions. Full statistics are reported in Appendix~\ref{app:mbe-significance}, Table~\ref{tab:mbe-significance-tests}.

\paragraph{Interpretation.}
The lack of a significant retrieval-condition difference contrasts with the larger gains on document-scoped FAQ (Section~\ref{sec:rag-courtlistener}), where improvements are strongest for temporal order, causality, and enablement. Experiment~A instead yields nearly identical chunked-text and EKG performance. EKG benefits therefore depend on the information required by the task rather than holding uniformly across legal QA.

\subsection{Experiment B: Open-Retrieval and Oracle-Document Analysis}
\label{sec:mbe-expB}

\paragraph{Setup.}
We evaluate 100 MBE-style multiple-choice questions from the \texttt{legal\_bench} subset, retrieving from the \texttt{reglab/barexam\_qa} corpus of 856,835 legal passages. Claude Sonnet~4.5 selects the final answer (A--D). We first measure open-retrieval Recall@10 and then downstream multiple-choice accuracy. BM25 and E5-large-v2 search the full corpus directly; hybrid Graph-RAG reranks the BM25 top-1,000 candidate set using graph-relevance scores where available.

\paragraph{Retrieval results.}
BM25 and E5 retrieve the gold passage in the top ten for 1\% of questions, while hybrid Graph-RAG reaches 8\% Recall@10 (Table~\ref{tab:barexam-retrieval}). The gold passage appears in the BM25 top-1,000 pool for eight questions, and graph reranking moves all eight into the final top ten. The Recall@10 gain therefore occurs when first-stage retrieval has already placed the relevant passage in the candidate pool; graph reranking cannot recover passages absent from that pool.

\begin{table}[t]
\centering
\begin{tabular}{lc}
\toprule
\textbf{Retrieval method} & \textbf{Recall@10} \\
\midrule
BM25                     & 0.0100 \\
E5-large-v2              & 0.0100 \\
\textbf{Hybrid Graph-RAG} & \textbf{0.0800} \\
\bottomrule
\end{tabular}
\caption{Open-retrieval performance on the 100-question Bar Exam QA subset.}
\label{tab:barexam-retrieval}
\end{table}

\paragraph{Downstream accuracy.}
BM25-RAG achieves 73\% accuracy and open Graph-RAG 72\%, both below the 76\% no-passage baseline (Table~\ref{tab:barexam-mcq}). The gold-passage condition reaches 79\%, showing that relevant evidence improves answer selection. In the open setting, however, the gold passage is absent from most retrieved contexts, so first-stage retrieval errors limit the downstream value of graph reranking.

\paragraph{Oracle-document setting.}
We additionally evaluate Graph-RAG when the relevant source document is known. The original configuration reaches 75\% accuracy but produces empty graph retrievals for 18 questions. A revised configuration adds document-scoped retrieval, an MCQ-specific prompt, and hybrid context combining the gold passage with graph evidence, reaching 82\%. Because these changes are introduced together, the 82\% result is a configuration-level comparison rather than a graph-only ablation.

\paragraph{Interpretation.}
These experiments separate first-stage retrieval from downstream use of structured evidence. Graph reranking raises open Recall@10 from 1\% to 8\%, but relevant passages enter the BM25 top-1,000 candidate pool for only eight of 100 questions, limiting downstream gains. In the oracle-document setting, the revised configuration reaches 82\%, but it changes retrieval scope, prompting, and context together. The results support EKG structure for organizing and reasoning over already-relevant evidence rather than replacing first-stage full-corpus retrieval.

\section{Conclusion}

We presented ARGUS, a staged pipeline for converting U.S.\ employment-discrimination complaints into source-grounded Event Knowledge Graphs. The strongest gains occur when tasks exploit participant, temporal, or causal structure: the graph-structured classifier achieves higher held-out Micro-F1 than raw and linearized baselines, and EKG-only document-scoped retrieval improves FAQ answer quality. The bar-exam experiments provide a boundary condition: Experiment~A shows no significant retrieval-condition difference, while open-retrieval gains occur only when relevant passages enter the first-stage candidate pool. ARGUS therefore supports organizing and reasoning over relevant legal evidence rather than replacing first-stage retrieval.

\section*{Limitations}
The corpus contains 153 U.S.\ federal employment-discrimination complaints from five districts and does not establish generalization to other claims, jurisdictions, or legal genres. Several evaluations use small samples: 24 chunks for mini-graph quality, five documents for the merger comparison, and 100--117 questions for the bar-exam studies. The FAQ benchmark is document-scoped, and the clustering analysis has no attorney relevance judgments, so neither establishes end-to-end similar-case retrieval. Human graph evaluation was performed by four project members rather than legal practitioners, and LLM judges may share biases with the systems they evaluate. OCR, extraction, coreference, and inferred temporal or causal errors can propagate. Finally, the configuration yielding the 82\% oracle-document result changes retrieval scope, prompting, and context together and should not be interpreted as a graph-only ablation.

\section*{Ethical Considerations}
Court complaints are public records but contain names and sensitive allegations. Automatically extracted roles, events, and causal relations can be incomplete or wrong and should not be treated as legal findings or used without attorney review. Any artifact release should follow the source repository's terms, minimize unnecessary personal information, and state clearly that ARGUS is a research prototype for retrieval and organization rather than legal advice.

\section*{Acknowledgments}

This work grew out of a master's capstone project at the University of Massachusetts Amherst conducted in collaboration with Bloomberg mentors. The authors
gratefully acknowledge the long-standing research and educational collaboration between Bloomberg and UMass Amherst, including student mentorship, joint
research, and scholarly exchange in natural language processing, data science, and related areas.

\bibliography{custom}

\appendix

% =========================================================
% APPENDIX A
% =========================================================

\section{Additional Implementation Figures}
\label{app:pipeline-figures}

\begin{figure}[t]
    \centering
    \includegraphics[width=\linewidth]{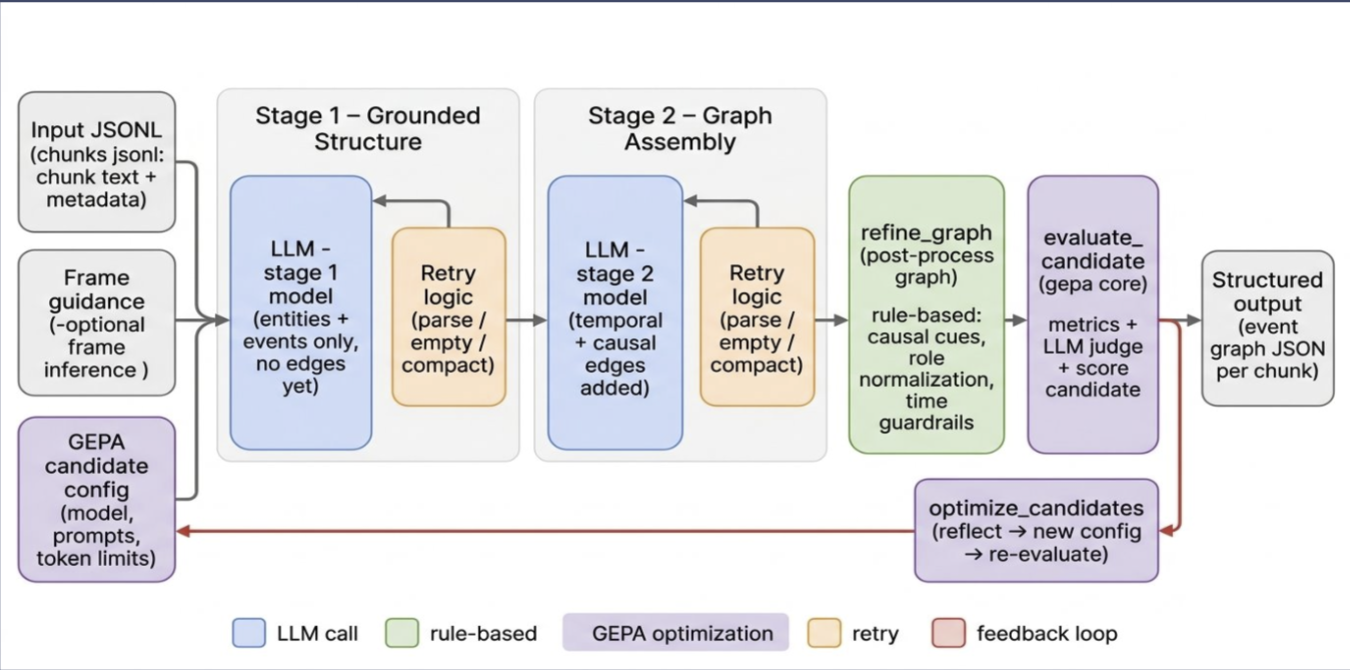}
    \caption{Two-stage mini-graph extractor with GEPA-driven optimization. Stage~1 extracts grounded entities and events; Stage~2 adds temporal and causal relations.}
    \label{fig:minigraph-extractor}
\end{figure}

\begin{figure}[t]
    \centering
    \includegraphics[width=\linewidth]{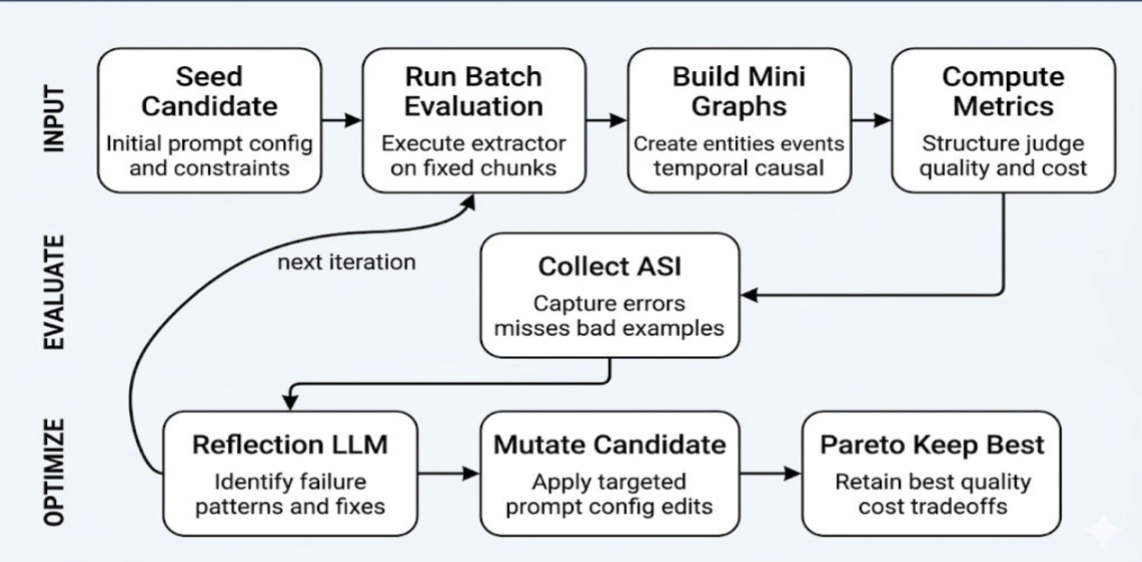}
    \caption{GEPA candidate optimization loop: seed, evaluate, reflect, mutate, and retain Pareto-optimal configurations.}
    \label{fig:gepa-loop}
\end{figure}

\begin{figure}[t]
    \centering
    \includegraphics[width=0.9\linewidth]{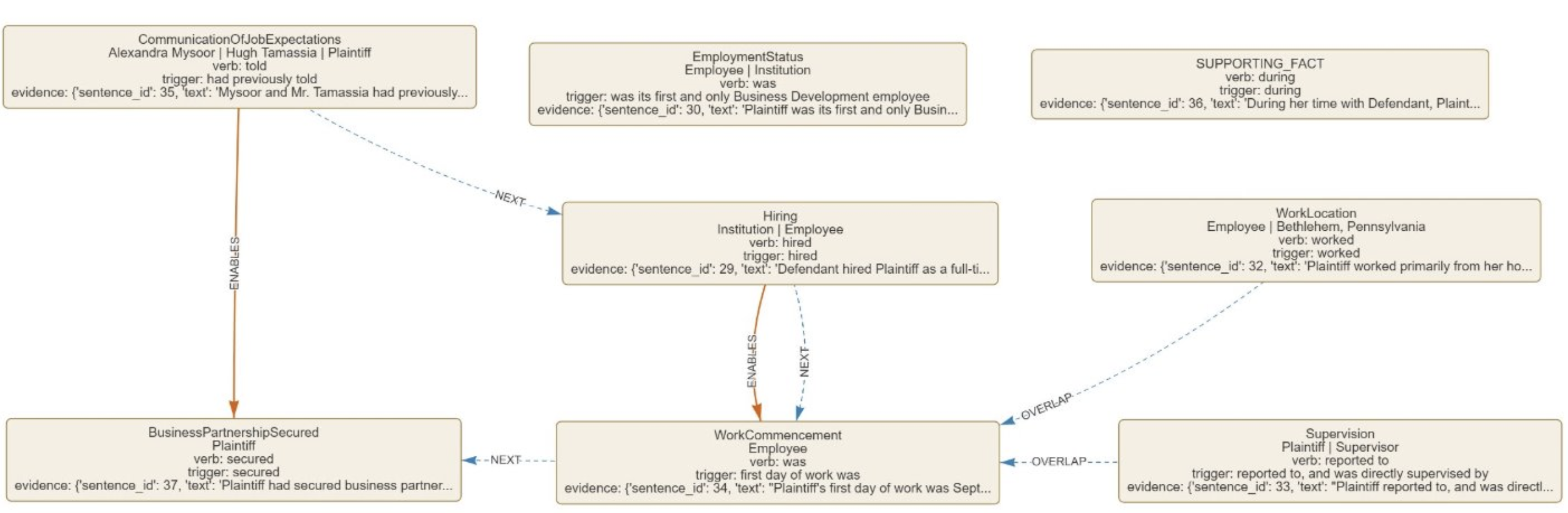}
    \caption{Example of an event mini-graph extracted from a single chunk.}
    \label{fig:minigraph-example}
\end{figure}

Figure~\ref{fig:minigraph-extractor} illustrates the two-stage mini-graph extraction procedure, Figure~\ref{fig:gepa-loop} summarizes the GEPA optimization loop, and Figure~\ref{fig:minigraph-example} shows an example extracted mini-graph.

% SUGGETION: The main experimental text currently refers to an
% annotation-dashboard figure (fig:annotation_dashboard), but no such
% figure is present in the current Appendix. Add the figure here if
% available, or remove that reference from the main text.
%
% SUGGETION: For full reproducibility, consider adding the exact semantic
% chunking configuration and the final optimized mini-graph extraction
% configuration used to generate the downstream EKGs. These details are
% not currently present in the Appendix.

% =========================================================
% APPENDIX B
% =========================================================

\section{Additional Experimental Details}
\label{app:experimental-details}

\subsection{One-Shot Event-Graph Prompt}
\label{app:oneshot-prompt}

The document-level baseline receives the complete sentence-labeled FACTS narrative and generates one event graph directly, without semantic chunking, intermediate mini-graphs, or staged document-level merging. The exact prompt is:

\begin{quote}
%\small\ttfamily
``You are a legal event extraction analyst. You are given one full legal document with sentence-labeled facts. Extract ONE overall document-level event graph, just extract what you think are events. Return strict JSON only. Schema requires: entities, events, temporal\_edges, causal\_edges with consistent IDs (E1..En, EV1..EVn). Events are the only nodes; entities attach to events as participants. Temporal relations: BEFORE, SAME\_TIME. Causal relation: CAUSES, ENABLES.''
\end{quote}

\subsection{Intra-Model Judge Stability}
\label{app:intra-stability}

We additionally evaluate the stability of document-level graph judgments under temperature variation. Table~\ref{tab:intra_model} reports the fraction of judgments that remain unchanged and the corresponding flip rate.

\begin{table}[t]
\centering
%\small
\begin{tabular}{lccc}
\toprule
\textbf{Model} & \textbf{Stability} & \textbf{Flip} & \textbf{AC1} \\
\midrule
DeepSeek & 80\%  & 20\% & 1.0 \\
Qwen     & 100\% & 0\%  & 1.0 \\
Mistral  & 100\% & 0\%  & 1.0 \\
\bottomrule
\end{tabular}
\caption{Intra-model stability under temperature variation.}
\label{tab:intra_model}
\end{table}

\subsection{Full GNN Architecture}
\label{app:gnn-details}

The EKG-GNN represents each merged document-level EKG as a homogeneous, undirected graph that contains both entity and event nodes and includes self-loops. Event--entity edges connect events to participant entities, while event--event edges represent temporal and causal relations. LEGAL-BERT encodes each node's short natural-language description. The resulting \texttt{[CLS]} embedding is projected to 256 dimensions and passed through two mean-neighbor graph-convolution layers. Each layer computes

\[
\begin{aligned}
h_i^{(l+1)} = \mathrm{ReLU}\!\Bigg(
    W_s h_i^{(l)}
    + W_n \frac{1}{|\mathcal{N}(i)|}
    \sum_{j\in\mathcal{N}(i)} h_j^{(l)}
\Bigg).
\end{aligned}
\]

Each layer is followed by dropout ($p=0.1$). Self-loops are stored in the graph but excluded from the neighbor average because the self contribution is handled separately by $W_s$. After two layers, node representations are mean-pooled and passed to a 23-label linear classifier.

The reported \texttt{ft-BERT} configuration jointly fine-tunes LEGAL-BERT and the graph layers using weighted binary cross-entropy and AdamW with a learning rate of $3\times10^{-4}$ and a weight decay of $0.01$. Training uses eight epochs, a batch size of 2, and gradient clipping at 1.0.

\subsection{Fact Extraction Details}
\label{app:fact-extraction}

Table~\ref{tab:us_holdout} reports the U.S.\ holdout F1 scores for classifier, hybrid, and LLM extraction modes.

\begin{table}[t]
\centering
%\small
\begin{tabular}{lcc}
\toprule
\textbf{Mode} & \textbf{LEGAL-BERT} & \textbf{LexLM} \\
\midrule
Classifier & 0.8718 & 0.8616 \\
Hybrid     & 0.8711 & 0.8627 \\
LLM        & 0.7115 & 0.7186 \\
\bottomrule
\end{tabular}
\caption{U.S.\ holdout Fact-class F1 by fact-extraction mode.}
\label{tab:us_holdout}
\end{table}

% SUGGETION: Confirm the U.S. holdout proportion before submission.
% The previous manuscript reports a 20\% holdout, whereas the uploaded
% implementation also contains a 60/40 train/holdout configuration.
% Once confirmed, state the correct split here and in the main text.

% =========================================================
% APPENDIX C
% =========================================================

\section{Detailed Clustering Analysis}
\label{app:clustering-details}

The clustering experiments are exploratory and examine whether graph-derived document representations yield coherent groupings of complaints. Table~\ref{tab:clustering} compares the principal feature configurations using silhouette score.

\begin{table}[t]
\centering
%\small
\begin{tabular}{lc}
\toprule
\textbf{Feature set} & \textbf{Silhouette} \\
\midrule
TF-IDF baseline & 0.091 \\
EKG full (2k dims) & 0.363 \\
EKG Legacy7D & 0.465 \\
\textbf{EKG Core6D ($K=4$)} & \textbf{0.649} \\
\bottomrule
\end{tabular}
\caption{Clustering silhouette by feature configuration.}
\label{tab:clustering}
\end{table}

We further evaluate three Core6D-based feature configurations (Table~\ref{tab:cluster-hybrid}), whose components are summarized in Table~\ref{tab:cluster-features}. Validation compares distance to the assigned centroid with an LLM-judged cluster-membership score at $K=4$.

Run A combines SALI claim tags with Core6D and produces one dominant cluster (111/14/6/2), with a weak distance--membership association ($\text{Pearson}\approx-0.06$, $n\approx17$). Run B combines Core6D with a 16-dimensional PCA-compressed EKG-context embedding and produces more balanced clusters, with $\text{Pearson}\approx-0.28$ and $\text{Spearman}\approx-0.31$ ($n=40$). Run C adds SALI tags and a $k$NN cross-document graph-neighbor component to Run B; it produces cluster sizes of 31/61/19/22 and correlations of $\text{Pearson}\approx-0.29$ and $\text{Spearman}\approx-0.37$ ($n=40$). The difference between Runs B and C is small, so this exploratory analysis does not establish a clear advantage for the additional Run C features.

Separately, the distance--membership correlation for Run B remains similar when the validation sample is increased from five to ten documents per cluster ($n=20$ versus $n=40$), providing a check on sensitivity to validation-sample size.

\begin{table}[t]
\centering
%\small
\setlength{\tabcolsep}{4pt}
\resizebox{\columnwidth}{!}{%
\begin{tabular}{clccc}
\toprule
\textbf{Run} & \textbf{Representation} & \textbf{Sizes ($K=4$)} & \textbf{Pearson} & \textbf{Spearman} \\
\midrule
A & SALI(23) + Core6D & 111/14/6/2 & $\approx-0.06$ & -- \\[2pt]
B & Core6D + Hybrid PCA(16) & balanced, 30s--50s & $\approx-0.28$ & $\approx-0.31$ \\[2pt]
C & A + Hybrid PCA + $k$NN(12) & 31/61/19/22 & $\approx-0.29$ & $\approx-0.37$ \\
\bottomrule
\end{tabular}}
\caption{Cluster validation across three feature configurations. Correlations compare centroid distance with LLM-judged cluster membership; more negative values indicate stronger association between proximity to the centroid and judged membership. Validation $n=40$ for Runs B and C and $n\approx17$ for Run A.}
\label{tab:cluster-hybrid}
\end{table}

\begin{table}[t]
\centering
%\small
\begin{tabular}{lccc}
\toprule
\textbf{Feature} & \textbf{A} & \textbf{B} & \textbf{C} \\
\midrule
SALI claim tags          & yes & no  & yes \\
Core6D structure         & yes & yes & yes \\
Hybrid PCA context       & no  & yes & yes \\
Graph neighbors ($k$NN)  & no  & no  & yes \\
\bottomrule
\end{tabular}
\caption{Features included in each clustering configuration.}
\label{tab:cluster-features}
\end{table}

\begin{figure}[t]
    \centering
    \includegraphics[width=0.9\linewidth]{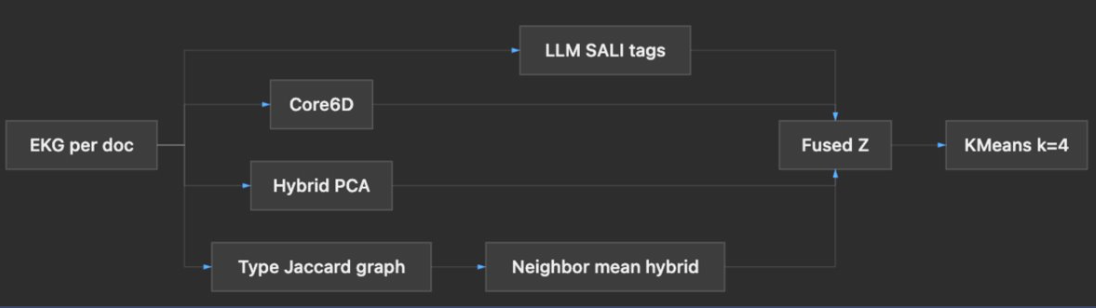}
    \caption{Run C fused representation: Core6D, hybrid PCA context, a type-Jaccard graph-neighbor component, and LLM-assigned SALI tags are combined into a single vector $Z$ before $K$-means clustering ($K=4$).}
    \label{fig:cluster-fusion}
\end{figure}

\begin{figure}[t]
    \centering
    \includegraphics[width=0.9\linewidth]{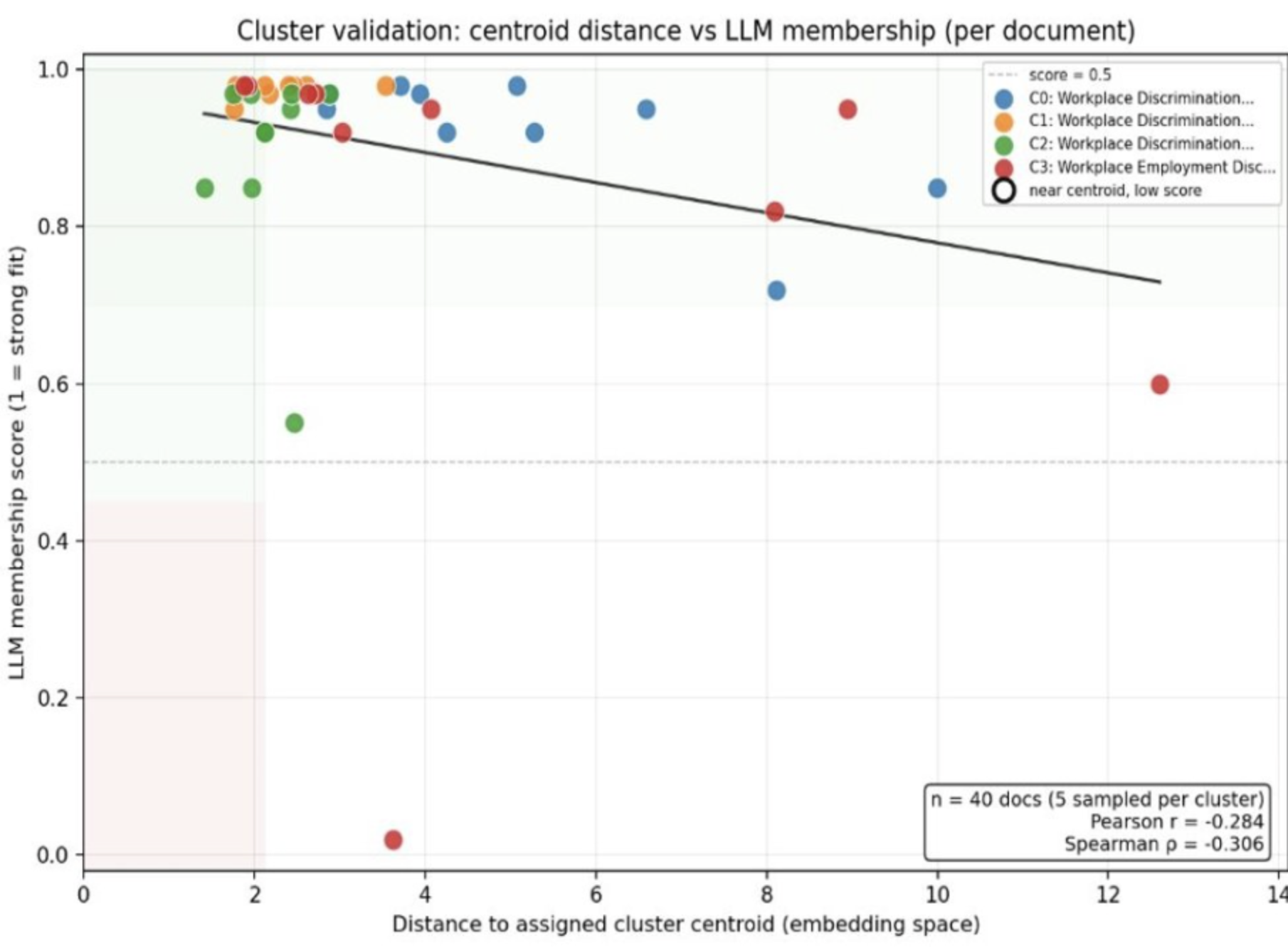}
    \caption{Run C validation: centroid distance versus LLM-judged cluster membership ($n=40$, Pearson $r\approx-0.29$, Spearman $\rho\approx-0.37$).}
    \label{fig:cluster-runc-scatter}
\end{figure}

\begin{figure}[t]
\centering
\begin{minipage}{0.48\linewidth}
    \centering
    \includegraphics[width=\linewidth]{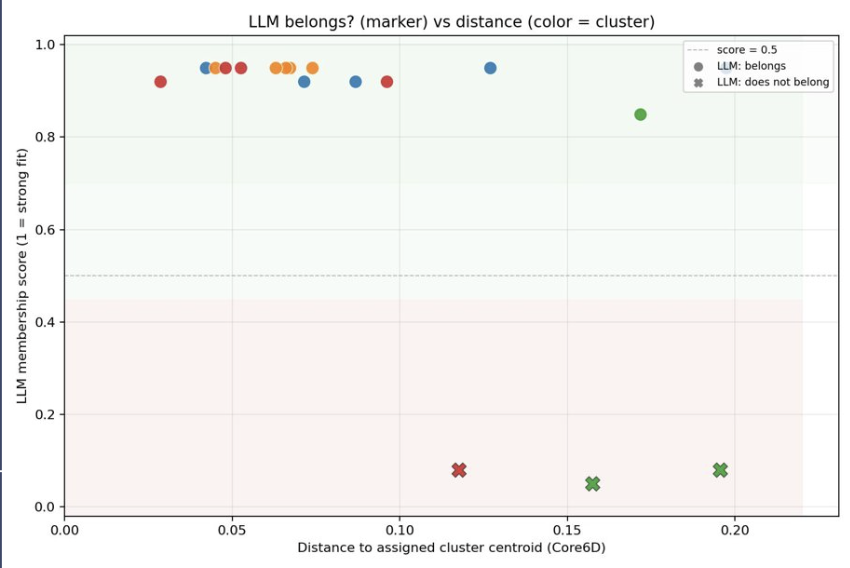}
    \caption*{Core6D only (baseline)}
\end{minipage}
\hfill
\begin{minipage}{0.48\linewidth}
    \centering
    \includegraphics[width=\linewidth]{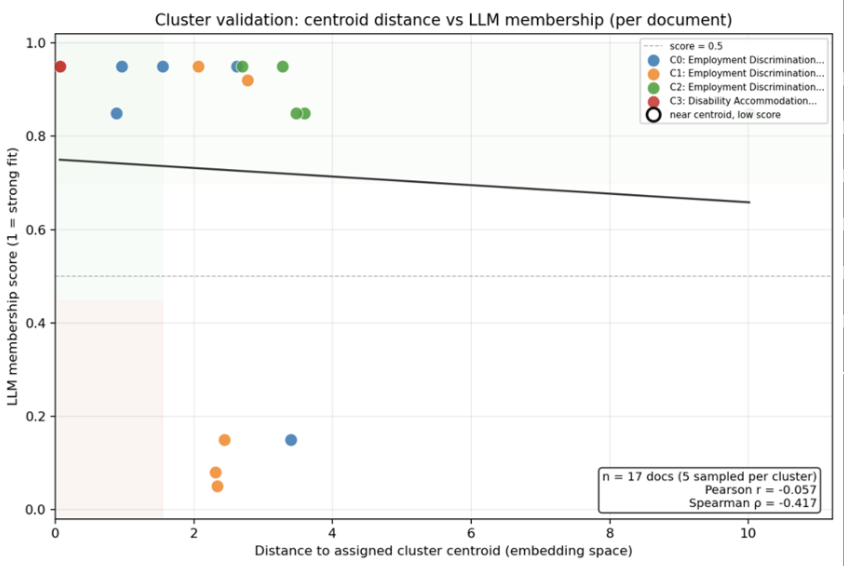}
    \caption*{Run A: SALI + Core6D}
\end{minipage}
\caption{Cluster validation by centroid distance and LLM-judged cluster membership for Core6D and Run A. This pairwise analysis preceded the three-configuration comparison in Table~\ref{tab:cluster-hybrid}.}
\label{fig:cluster-scatter}
\end{figure}

% =========================================================
% APPENDIX D
% =========================================================

\section{Additional Evaluation Tables}
\label{app:additional-evaluation}

\subsection{FAQ Evaluation}
\label{app:faq-significance}

Table~\ref{tab:rag-by-type} reports document-scoped FAQ performance by question type.

\begin{table}[t]
\centering
%\small
\begin{tabular}{lcc}
\toprule
\textbf{Question type} & \textbf{Raw F1} & \textbf{EKG F1} \\
\midrule
Temporal order     & 0.381 & 0.538 \\
Causal/enablement  & 0.387 & 0.522 \\
Temporal overlap   & 0.241 & 0.384 \\
Party resolution   & 0.152 & 0.197 \\
\bottomrule
\end{tabular}
\caption{FAQ performance by question type.}
\label{tab:rag-by-type}
\end{table}

For the $n=299$ questions successfully scored under all three retrieval conditions, we compare methods using paired bootstrap confidence intervals, Wilcoxon signed-rank tests, and sign-flip paired permutation tests. Table~\ref{tab:faq-sig} reports the paired differences.

\begin{table}[t]
\centering
%\small
\resizebox{\columnwidth}{!}{%
\begin{tabular}{lccc}
\toprule
\textbf{Comparison} & \textbf{$\Delta$F1} & \textbf{Bootstrap 95\% CI} & \textbf{Wilcoxon $p$} \\
\midrule
EKG $-$ Raw    & +0.123 & [+0.109, +0.137] & $3.4\times10^{-39}$ \\
EKG $-$ Hybrid & +0.083 & [+0.069, +0.096] & $2.3\times10^{-27}$ \\
Hybrid $-$ Raw & +0.040 & [+0.030, +0.052] & $8.5\times10^{-11}$ \\
\bottomrule
\end{tabular}}
\caption{Paired significance tests for the FAQ benchmark ($n=299$). All comparisons are significant at $\alpha=0.05$; sign-flip permutation tests agree in every comparison ($p<10^{-3}$, not shown).}
\label{tab:faq-sig}
\end{table}

Table~\ref{tab:faq-sig-bytype} reports the EKG-versus-raw comparison separately by question type. All categories with at least 12 examples show a significant difference; the three-example temporal-relation category is too small to support a conclusion.

\begin{table}[t]
\centering
%\small
\resizebox{\columnwidth}{!}{%
\begin{tabular}{lccc}
\toprule
\textbf{Question type} & \textbf{$n$} & \textbf{$\Delta$F1 (EKG$-$Raw)} & \textbf{Wilcoxon $p$} \\
\midrule
Temporal order     & 119 & +0.157 & $1.0\times10^{-18}$ \\
Causal/enablement  & 99  & +0.134 & $2.7\times10^{-16}$ \\
Temporal overlap   & 12  & +0.143 & $2.4\times10^{-3}$ \\
Party resolution   & 66  & +0.045 & $8.0\times10^{-5}$ \\
Temporal relation  & 3   & +0.019 & 0.25 \\
\bottomrule
\end{tabular}}
\caption{EKG-versus-raw FAQ results by question type. Categories with $n\geq12$ are significant at $\alpha=0.05$; no conclusion is drawn for the temporal-relation category ($n=3$).}
\label{tab:faq-sig-bytype}
\end{table}

\subsection{Bar-Exam Experiment A}
\label{app:mbe-significance}

Table~\ref{tab:mbe-retrieval-accuracy} reports multiple-choice accuracy under the retrieval conditions used in Experiment~A.

\begin{table}[t]
\centering
%\small
\begin{tabular}{lc}
\toprule
\textbf{Retrieval condition} & \textbf{MC accuracy} \\
\midrule
Raw document (top 3)        & 0.8547 \\
Chunked text (top 3)        & 0.8632 \\
EKG graph (top 3)           & 0.8632 \\
Gold EKG subset ($n=100$)   & 0.8600 \\
\bottomrule
\end{tabular}
\caption{MBE multiple-choice accuracy by retrieval condition in Experiment~A ($n=117$, Claude Sonnet~5).}
\label{tab:mbe-retrieval-accuracy}
\end{table}

Pairwise significance is evaluated using exact McNemar tests, paired bootstrap confidence intervals, and Wilcoxon signed-rank tests. No pairwise comparison reaches statistical significance (Table~\ref{tab:mbe-significance-tests}).

\begin{table}[t]
\centering
\scriptsize
\setlength{\tabcolsep}{3pt}
\resizebox{\columnwidth}{!}{%
\begin{tabular}{lccc}
\toprule
\textbf{Comparison} & \textbf{McNemar $p$} & \textbf{Bootstrap 95\% CI} & \textbf{Wilcoxon $p$} \\
\midrule
Document vs.\ chunk & 1.000 & $[-5.13,+3.42]$ pp & 0.813 \\
Document vs.\ EKG   & 1.000 & $[-5.98,+4.27]$ pp & 0.820 \\
Chunk vs.\ EKG      & 0.688 & $[-4.27,+4.27]$ pp & 1.000 \\
\bottomrule
\end{tabular}}
\caption{Pairwise significance tests for MBE accuracy in Experiment~A. No comparison is significant at $\alpha=0.05$.}
\label{tab:mbe-significance-tests}
\end{table}

\subsection{Bar-Exam Experiment B}
\label{app:mbe-experiment-b}

Table~\ref{tab:barexam-retrieval} reports Recall@10 for the full-corpus retrieval conditions in Experiment~B.

Table~\ref{tab:barexam-mcq} reports downstream multiple-choice accuracy under open-retrieval, gold-passage, and oracle-document conditions.

\begin{table}[t]
\centering
%\small
\resizebox{\columnwidth}{!}{%
\begin{tabular}{lc}
\toprule
\textbf{Context condition} & \textbf{MCQ accuracy} \\
\midrule
No passage                                  & 0.7600 \\
BM25-RAG (open retrieval, top 10)           & 0.7300 \\
Hybrid Graph-RAG (open retrieval, top 10)   & 0.7200 \\
Gold passage                                & 0.7900 \\
Original oracle-document Graph-RAG          & 0.7500 \\
\textbf{Improved oracle-document Graph-RAG} & \textbf{0.8200} \\
\bottomrule
\end{tabular}}
\caption{Downstream Bar Exam multiple-choice accuracy in Experiment~B (Claude Sonnet~4.5).}
\label{tab:barexam-mcq}
\end{table}

\end{document}